\documentclass[conference]{IEEEtran}

\usepackage{cite}
\usepackage{amsmath,amssymb,amsfonts,mathtools,bm}
\usepackage{algorithm}
\usepackage{algpseudocode}
\usepackage{graphicx}
\usepackage{textcomp}
\usepackage{xcolor}
\usepackage{booktabs}
\usepackage{array}
\usepackage{tikz}
\usetikzlibrary{arrows.meta,positioning,fit,backgrounds,calc}
\usepackage{url}
\usepackage{amsthm}

\IEEEsetsidemargin{c}{0pt}
\begin{document}
	
	\title{Weighted-AoI-Driven UAV Trajectory and Reporting for Dynamic CKM Tracking}
	
	\author{
		\IEEEauthorblockN{Zeyu Zhao$^{1}$, Yueling Che$^{2}$ and Sheng Luo$^{3}$}
		\IEEEauthorblockA{College of Computer Science and Software Engineering, Shenzhen University, Shenzhen, China\\
			2100271103@email.szu.edu.cn$^{1}$, yuelingche@szu.edu.cn$^{2}$,
			sluo@szu.edu.cn$^{3}$}}
	
	\maketitle
	
	\begin{abstract}
		Existing channel knowledge map (CKM) construction assumes quasi-static environments, failing when channels rapidly decorrelate. This paper investigates continuous online dynamic CKM tracking over an airborne 2D plane via a UAV fleet. We formulate a long-term error minimization problem under UAV mobility and reporting constraints. This problem, however, is intractable with unobservable errors for unmeasured cells. To tackle this, we propose a metric-driven scheduling approach featuring the Weighted-Age of Information (W-AoI). W-AoI evaluates cells by combining exact temporal staleness with spatial uncertainty predicted by a 2D U-Net. We prove that top-$K$ W-AoI selection maximizes the one-slot error reduction bound. Guided by this guarantee, we develop a lightweight greedy trajectory policy incorporating continuous along-path sensing. Simulations demonstrate our method reduces steady-state error by 3.8 dB$^2$ and 4.8 dB$^2$ over AoI-only and FIFO baselines, closely approaching an idealized oracle.
	\end{abstract}
	
	
	\section{Introduction}
	
	As unmanned aerial vehicles (UAVs) increasingly integrate into cellular networks \cite{3GPPtr36}, channel knowledge maps (CKMs) have become a vital tool for environment-aware communications \cite{zeng2024tutorial}. By providing location-specific channel priors such as expected signal-to-interference-plus-noise ratio (SINR), a CKM enables proactive trajectory planning and resource allocation rather than reactive outage handling \cite{zhang2020radio}. Since ground drive tests cannot reach the airspace and ray tracing demands exact 3D physical models, UAVs serve as natural mobile collectors. However, constrained by flight speed, inter-UAV safety distance, and uplink reporting budgets, a UAV fleet can only measure and report a small fraction of the target area per slot, leaving the server with sparse observations.
	
	To recover a complete map from sparse samples, existing studies rely on spatial interpolation techniques ranging from Kriging \cite{sato2017kriging} and deep autoencoders \cite{teganya2021deep} to generative diffusion models \cite{rong20253dckmdiff}. From the data collection perspective, UAV trajectory policies have been designed for active spectrum surveying \cite{shrestha2022spectrum} and partial-CKM navigation \cite{song2026trajectory}, alongside theoretical analysis on the spatial sampling density required for target accuracy \cite{xu2024much}. While effective against spatial sparsity, most existing approaches rely on a \emph{quasi-static} assumption: they treat channel statistics as time-invariant during the survey and pool measurements from different slots into a single static map, overlooking short-term temporal variations.
	
	In practical urban networks, dynamic interference, traffic fluctuations, and moving scatterers cause channels to decorrelate before UAVs can sweep the entire grid, necessitating continuous online updates for dynamic CKM maintenance. Although recent works have explored dynamic radio map tracking via statistical hypothesis testing \cite{katagiri2020dynamic} and adaptive channel state information (CSI) map path planning via deep reinforcement learning (DRL) \cite{zhu2021path}, key challenges remain unresolved. First, fixed-sensor updating rules \cite{katagiri2020dynamic} do not account for UAV mobility and collision-avoidance constraints. Second, Age-of-Information (AoI)-based scheduling \cite{yates2021age} tracks temporal staleness but treats all spatial cells equally, ignoring that regions with sharp channel variations degrade faster. Third, DRL trajectory planners \cite{zhu2021path} require heavy offline training without explicit theoretical bounds on estimation error.
	
	In this paper, we investigate the maintenance of a dynamic CKM over an aerial plane using a UAV fleet, and formulate a long-term CKM estimation error minimization problem under practical UAV operational constraints. Since the true error of unmeasured cells is unobservable, we employ a two-phase 2D U-Net at the server to predict both the current CKM and an Absolute Error Map (AEM) from historical sparse measurements and AoI states. Combining the exact AoI with the predicted AEM, we propose the \emph{Weighted-AoI (W-AoI)} metric and prove that it upper-bounds the true cell error, with Top-$K$ W-AoI selection maximizing the one-slot bound reduction under idealized conditions. Guided by this guarantee, we develop a lightweight greedy trajectory policy where each UAV navigates within a Voronoi subregion toward the highest-W-AoI reachable cell while continuously measuring cells along its flight path. Finally, extensive simulations demonstrate that our method significantly outperforms baseline approaches and closely approaches an idealized oracle.
	
	\textit{Notation}: $|\mathcal{X}|$ is set size, $\odot$ element-wise product, $\|\cdot\|_2$
	the Euclidean norm; bold letters are vectors or matrices.
	
	\section{System Model}
	\label{sec:sys}
	
	\subsection{Network and Channel Model}
	Consider a low-altitude target area modeled as a 2D horizontal plane $\mathcal{D} = [0, L]^2$ at a fixed altitude $h_0$. A set of $M$ ground base stations (GBSs), denoted by $\mathcal{M} = \{1, \dots, M\}$ with locations $\boldsymbol q_m = [x_m, y_m, z_m]^T$, provides downlink coverage over the aerial plane. The region $\mathcal{D}$ is discretized into an $I \times J$ grid of square cells, where $I = J = L/\Delta_D$. The grid resolution $\Delta_D$ is small such that the channel quality is assumed homogeneous within each cell. We denote $\boldsymbol q \in \mathcal{D}$ as the center coordinate of a generic cell. For brevity, $\mathcal D$ also denotes the discrete set of $IJ$ cell centers in what follows. To construct the CKM, a fleet of $U$ rotary-wing UAVs, denoted by $\mathcal{U} = \{1, \dots, U\}$, acts as mobile sensors navigating through $\mathcal{D}$ to measure the spatial-temporal channel states and then report to their serving GBSs. The GBSs are connected via fiber backhaul to a central server (e.g., GBS 1), which aggregates all UAV measurements to maintain the global CKM. The system operates in discrete time slots $\mathcal{T} = \{1, \dots, T\}$ with duration $\Delta t$.
	
	Let $G_m(\boldsymbol q, t)$ be the channel gain from GBS $m$ to a typical aerial receiver located at cell $\boldsymbol q$ in slot $t$. The gain is composed of large-scale and small-scale fading components:
	\begin{equation} \label{eq:gain}
		G_m(\boldsymbol q, t) = \bar{G}_m(\boldsymbol q) |h_m(\boldsymbol q, t)|^2,
	\end{equation}
	where $\bar{G}_m(\boldsymbol q) = 10^{[G_a(\theta_m(\boldsymbol q)) - \text{PL}_m(\boldsymbol q) - S_m(\boldsymbol q)]/10}$ captures the quasi-static large-scale channel gain, including the vertical antenna array gain $G_a$, the 3GPP urban-micro path loss $\text{PL}_m$ \cite{3GPPtr36}, and the spatial correlated shadowing $S_m(\boldsymbol q)$. The term $h_m(\boldsymbol q, t)$ represents the dynamic small-scale fading, modeled as a first-order Gauss-Markov process:
	\begin{equation} \label{eq:fading}
		h_m(\boldsymbol q, t) = \rho_h h_m(\boldsymbol q, t-1) + \sqrt{1 - \rho_h^2} e_m(\boldsymbol q, t),
	\end{equation}
	with $e_m \sim \mathcal{CN}(0, 1)$ and correlation coefficient $\rho_h \in [0, 1]$.

	All GBSs share one frequency band. Let $\alpha_m(t)\in\{0,1\}$ be one when GBS $m$
	transmits in slot $t$ and zero otherwise; $\alpha_m(t)$ follows a two-state Markov chain with
	stationary probability $l_m=\mathbb{E}[\alpha_m(t)]$ and correlation $\rho_\alpha \in [0, 1]$,
	\begin{equation}
		\begin{aligned}
			\Pr\{\alpha_m(t)\!=\!1\mid\alpha_m(t{-}1)\!=\!0\}&=(1-\rho_\alpha)\,l_m,\\
			\Pr\{\alpha_m(t)\!=\!1\mid\alpha_m(t{-}1)\!=\!1\}&=\rho_\alpha+(1-\rho_\alpha)\,l_m,
		\end{aligned}
		\label{eq:markov}
	\end{equation}
	The activity process reduces to independent Bernoulli activity at $\rho_\alpha=0$; otherwise the interference a cell sees stays correlated over
	$1/(1-\rho_\alpha)$ slots. The instantaneous downlink SINR experienced at cell $\boldsymbol q$ from GBS $m$ is:
	\begin{equation} \label{eq:sinr}
		\gamma_m(\boldsymbol q, t) = \frac{P_m G_m(\boldsymbol q, t)}{\sum_{m' \neq m} \alpha_{m'}(t) P_{m'} G_{m'}(\boldsymbol q, t) + \sigma^2},
	\end{equation}
	where $P_m$ is the transmit power and $\sigma^2$ is the noise power.
	
	\subsection{Dynamic CKM Formulation}
	To avoid frequent handovers caused by fast fading, a cell $\boldsymbol q$ is associated with a serving GBS based on the expected large-scale SINR, given by $\bar{\gamma}_m(\boldsymbol q) = P_m \bar{G}_m(\boldsymbol q) / (\sum_{m' \neq m} l_{m'} P_{m'} \bar{G}_{m'}(\boldsymbol q) + \sigma^2)$. The index of the serving GBS is determined as $m^*(\boldsymbol q) = \arg\max_{m \in \mathcal{M}} \bar{\gamma}_m(\boldsymbol q)$ and is kept static.
	
	Existing CKMs predominantly focus on tracking quasi-static large-scale components, attempting to store the reference map $\bar{C}(\boldsymbol q) = 10 \log_{10} \bar{\gamma}_{m^*}(\boldsymbol q)$ (in dB) \cite{zhang2020radio} and \cite{song2026trajectory}. Since the numerator and the denominator of
	\eqref{eq:sinr} are independent and $1/x$ is convex, Jensen's inequality gives
	$\mathbb{E}[\gamma_{m^{*}}]\ge\bar\gamma_{m^{*}}$, hence
	$\bar C\le 10\log_{10}\mathbb{E}[\gamma_{m^{*}}]$ \cite{zhang2020radio}. The inequality implies $\bar{C}(\boldsymbol q)$ has a definite downward bias compared to the true mean. More importantly, because the bias varies from cell to cell (depending on local interference and scattering), the bias cannot be eliminated by simply shifting the entire map by a global constant. Thus, the gap $|C(\boldsymbol q, t) - \bar{C}(\boldsymbol q)|$ represents an inherent error that no quasi-static construction can remove.
	
	To precisely capture the variations, we define the \emph{dynamic CKM} as the instantaneous SINR map:
	\begin{equation}\label{eq:ckm}
		C(\boldsymbol q, t) = 10 \log_{10} \gamma_{m^*}(\boldsymbol q, t) \quad [\text{dB}].
	\end{equation}
	The network's objective is to estimate and track $C(\boldsymbol q, t)$ slot-by-slot, keeping $\bar{C}(\boldsymbol q)$ only as a reference baseline.
	
	\subsection{UAV Measurement and Age of Information}
	Let $\boldsymbol p_{u,t}\in\mathcal D$ be the horizontal position of UAV $u$ at the
	beginning of slot $t$. Subject to a maximum speed $V_{\max}$, the flight distance per slot is
	$\|\boldsymbol p_{u,t+1}-\boldsymbol p_{u,t}\|_2\le V_{\max}\Delta t$. During navigation, the UAV $u$ records the channel quality of the cells along the trajectory. The maximum number of cells crossed per slot is bounded by $n_{\max} = \lceil \sqrt{2} V_{\max} \Delta t / \Delta_D \rceil + 1$.
	
	Measurements are appended to an on-board finite queue $\mathcal{B}_u(t)$ of capacity $Q$, whose oldest entries are
	dropped once $|\mathcal B_u(t)|>Q$. In each slot, each UAV $u$ reports a subset of measurements
	\begin{equation}
		\mathcal S_{u,t}=\operatorname{Select}_K\big(\mathcal B_u(t)\big),\qquad
		|\mathcal S_{u,t}|\le K,
		\label{eq:report}
	\end{equation}
	where $\operatorname{Select}_K(\cdot)$ is the selection rule studied in
	Section~\ref{sec:solve}. The aggregated reports from the fleet is $\mathcal{S}_t = \bigcup_{u \in \mathcal{U}} \mathcal{S}_{u, t}$. 
	
	Because the fleet reports at most $KU \ll IJ$ measurements per slot, most cells remain unmeasured. 
	Moreover, since the dynamic channel decorrelates over time, only sparse measurements are available per slot. We track the staleness of the unmeasured cells using the AoI:
	\begin{equation}\label{eq:aoi}
		A(\boldsymbol q, t) = 
		\begin{cases}
			t - \tau(\boldsymbol q, t), & \boldsymbol q \in \mathcal{S}_t, \\
			\min\{A(\boldsymbol q, t-1) + 1, A_{\max}\}, & \text{otherwise},
		\end{cases}
	\end{equation}
	where $\tau(\boldsymbol q, t) < t$ is the generation slot of the measurement reported at time $t$ for cell $\boldsymbol q$, and $A_{\max}$ bounds the AoI. Since relying solely on sparse and outdated measurements inevitably degrades the CKM, to minimize the resultant construction error, the central server needs to jointly optimize the UAV trajectories and reporting decisions.
	
	\section{Problem Formulation and Learning-Based Prediction}
	\label{sec:prob}
	
	\subsection{Dynamic CKM Tracking Problem}
	To maintain an accurate dynamic CKM despite sparse measurements, the central server maintains a dynamically updated estimate, denoted by $C'(\boldsymbol q, t)$. The accuracy of the CKM estimate is evaluated by the spatial error distribution, defined as the AEM:
	$
	E(\boldsymbol q, t) = \vert{}C(\boldsymbol q, t) - C'(\boldsymbol q, t)\vert{}. $
	AEM $E(\boldsymbol q, t)$ must be distinguished from the static gap $\vert{}C(\boldsymbol q, t) - \bar{C}(\boldsymbol q)\vert{}$ discussed in Section II. While the static gap represents an unremovable inherent bias of existing static maps, $E(\boldsymbol q, t)$ quantifies how accurately the central server tracks the true dynamic channel $C(\boldsymbol q, t)$.
	
	To keep the tracking error as small as possible, the central server needs to jointly optimize the UAV trajectories $\boldsymbol Q = \{\boldsymbol p_{u, t}\}$ and the reporting decisions $\mathcal{S} = \{\mathcal{S}_{u, t}\}$, to minimize the time-averaged Mean Squared Error (MSE) over the entire area $\mathcal{D}$. The problem is formulated as:
	\begin{align}
		(\mathrm{P})\quad \min_{\boldsymbol Q,\,\mathcal S}\quad
		& \frac{1}{IJT}\sum_{t=1}^{T}\sum_{\boldsymbol q\in\mathcal D}E^{2}(\boldsymbol q,t)
		\nonumber\\
		\text{s.t.}\quad
		& |\mathcal S_{u,t}|\le K, \label{eq:c1}\\
		& |\mathcal B_u(t)|\le Q, \label{eq:c2}\\
		& \mathcal S_{u,t}\subseteq\mathcal B_u(t), \label{eq:c3}\\
		& \mathcal B_u(t)=\mathcal B_u(t-1)\cup\mathcal P_{u,t}, \label{eq:c4}\\
		& \|\boldsymbol p_{u,t+1}-\boldsymbol p_{u,t}\|_2\le V_{\max}\Delta t,\quad
		\boldsymbol p_{u,t}\in\mathcal D, \label{eq:c5}\\
		& \|\boldsymbol p_{u,t}-\boldsymbol p_{u',t}\|_2\ge d_{\min},\quad u\neq u',
		\label{eq:c6}
	\end{align}
	with \eqref{eq:c1}--\eqref{eq:c6} required for every $u\in\mathcal U$ and $t\in
	\mathcal T$. Specifically, \eqref{eq:c1} limits the maximum number of measurements a UAV can report per slot due to the uplink budget; \eqref{eq:c2} enforces the finite on-board queue capacity; \eqref{eq:c3} ensures only measurements currently in the queue can be reported; \eqref{eq:c4} governs the queue evolution, removing reported data and appending new UAV measurements; \eqref{eq:c5} bounds the flight distance per slot based on the maximum speed $V_{\max}$ while keeping the UAV within the target area; and \eqref{eq:c6} mandates a minimum safety distance $d_{\min}$ to prevent inter-UAV collisions.
	
	Problem (P) is a highly coupled, mixed-integer, and stochastic optimization problem. A fundamental paradox in solving (P) directly is that evaluating the true AEM $E(\boldsymbol q, t)$ in the objective requires the true dynamic map $C(\boldsymbol q, t)$. The dynamic map is inherently unobservable, because reconstructing the map is the very goal of the system.
	
	\subsection{Spatial--Temporal Prediction via 2-D U-Net}
	
	Without access to the true AEM, a common practice is to rely on the observable AoI, prioritizing cells that have not been measured for the longest time. However, AoI alone is an inadequate proxy for actual channel deviation. AoI merely measures the time elapsed since the last update; a cell with a large AoI in a stable environment may still hold an accurate value, whereas a cell with a small AoI under bursty interference might already be highly erroneous. 
	
	Therefore, rather than relying solely on AoI, predicting the actual spatial error distribution is imperative. We introduce two 2-D U-Nets to predict the estimated map $\hat{\boldsymbol C}(t)$ and the predicted AEM $\hat{\boldsymbol E}(t)$ respectively, at the beginning of slot $t$. The two predictors use 2-D U-Nets similar to
	the design in \cite{teganya2021deep}, with three down-sampling stages and
	$N_{\rm ne}=1.2$~M parameters each, but with independent weights. 
	
	The predictors utilize the previous map estimate and the label of the previous slot,
	$\boldsymbol C'(t-1)$ and $\boldsymbol Y(t-1)$, together with a mask of fresh
	measurements. Let $\boldsymbol I_t = \mathbf{1}\{A(\cdot, t) \leq a_{\text{th}}\}$ be a
	binary indicator for cells whose AoI is below a freshness threshold $a_{\text{th}}$.
	Since a cell with $A>a_{\rm th}$ was measured long ago and may mislead the network
	\cite{xu2024much}, the label of slot $t$ keeps the measured value only at fresh cells
	and the previous estimate elsewhere,
	\begin{equation}
		Y(\boldsymbol q,t)=
		\begin{cases}
			C(\boldsymbol q,\tau(\boldsymbol q,t)), & A(\boldsymbol q,t)\le a_{\rm th},\\
			C'(\boldsymbol q,t-1), & \text{otherwise}.
		\end{cases}
		\label{eq:label}
	\end{equation}
	The label is fixed once the reports of slot $t$ are in, so the same image serves two purposes: as the training target of slot $t$ and, at the start of slot $t+1$, as an input carrying the last reliable measurements. The input tensor to the networks is therefore
	$\boldsymbol X^{l}(t)=\big[\boldsymbol M^{l}(t),\ \boldsymbol I_{t-1},\
	\boldsymbol C'(t-1),\ \boldsymbol Y(t-1)\big]$,
	where $\boldsymbol Y(t-1)$ supplies the measurements that were fresh at $t-1$.
	Unlike $\boldsymbol C'(t-1)$, $\boldsymbol Y(t-1)$ distinguishes the measured values
	from predicted values. $\boldsymbol M^l(t)$ is a Bernoulli mask generated with
	probability $p=0.8$ (Table~\ref{tab:param}), used to randomly hide
	a fraction of fresh cells during training to force spatial interpolation (set to an
	all-ones mask at inference). The two predictors, $f^C$ and
	$f^E$, are defined as:
	\begin{equation}
		\hat{\boldsymbol C}^{l}(t)=f^{C}\big(\boldsymbol X^{l}(t)\big),\quad
		\hat{\boldsymbol E}^{l}(t)=f^{E}\big(\boldsymbol X^{l}(t),\hat{\boldsymbol C}^{l}(t)\big).
		\label{eq:predict}
	\end{equation}
	The error predictor $f^E$ takes the predicted map $\hat{\boldsymbol C}^{l}(t)$ as an additional input to evaluate the residual uncertainty.
	
	\subsection{Online Update and Training}
	Once the subset of measurements $\mathcal{S}_t$ is reported to the central server by the end of slot $t$, the global map is updated online. The reported cells adopt the actual measured values, while the unmeasured cells retain the predicted values:
	\begin{equation}
		C'(\boldsymbol q,t)=
		\begin{cases}
			C(\boldsymbol q,\tau(\boldsymbol q,t)), & \boldsymbol q\in\mathcal S_t,\\
			\hat C(\boldsymbol q,t), & \text{otherwise}.
		\end{cases}
		\label{eq:update}
	\end{equation}
	
	The neural networks are trained online to adapt to the time-varying channel statistics. Since stale cells ($A > a_{\text{th}}$) contain severely drifted values that mislead the learning process, the network is trained exclusively on fresh measurements, creating a self-supervised target $Y(\boldsymbol q, t)$ in (\refeq{eq:label}). The loss function $L^C$ for the map predictor $f^C$ comprises three components: map-estimation loss on visible fresh cells, interpolation loss on hidden fresh cells (masked by $\boldsymbol M^l$), and a temporal consistency loss on stale cells:
	\begin{multline}
		L^{C}=\frac{1}{N_{\rm ma}}\sum_{l=1}^{N_{\rm ma}}\Big[
		\tfrac{\|(\hat{\boldsymbol C}^{l}-\boldsymbol{Y})\odot\boldsymbol M^{l}\odot\boldsymbol I_t\|_2^{2}}{IJ}\\
		+\lambda_1\tfrac{\|(\hat{\boldsymbol C}^{l}-\boldsymbol{Y})\odot(1-\boldsymbol M^{l})\odot\boldsymbol I_t\|_2^{2}}{IJ}
		+\lambda_2\tfrac{\|(\hat{\boldsymbol C}^{l}-\boldsymbol{Y})\odot(1-\boldsymbol I_t)\|_2^{2}}{IJ}\Big],
		\label{eq:lossc}
	\end{multline}
	where $\odot$ denotes the element-wise product, $N_{\rm ma}$ is the number of masks per batch, and $\lambda_1, \lambda_2$ are weighting coefficients. The error predictor $f^E$ is trained similarly against the residual $|\boldsymbol{Y}(t) - \hat{\boldsymbol C}^{l}(t)|$ with gradients detached from $\hat{\boldsymbol C}^{l}$.
	
	With the predicted AEM $\hat{E}(\boldsymbol q, t)$ available, the unobservable true error $E(\boldsymbol q, t)$ in the objective of Problem (P) is substituted by the predicted AEM. Nevertheless, solving (P) remains highly intractable. The variables are combinatorial sets, the stochastic channel transitions lack a closed-form model, and the decisions are strictly coupled across time slots via the queue evolution in \eqref{eq:c4} and the UAV mobility in \eqref{eq:c5}. To handle the complexity, we design a greedy yet theoretically guaranteed metric-driven online algorithm in Section~\ref{sec:solve}.
	
	\section{Metric-Driven UAV Trajectory and Reporting}
	\label{sec:solve}
	
	In this section, we first theoretically derive the weighted-AoI (W-AoI) as a unified cell-level priority metric, and subsequently design a mobility-constrained greedy trajectory policy to actively suppress the W-AoI.
	
	\subsection{Weighted-AoI and Error Upper Bound}
	
	Intuitively, the longer a cell remains unmeasured, the more its true channel drifts; and this drift occurs faster in regions with severe spatial variations. Hence, the W-AoI is defined as the product of the AoI and the predicted AEM as follows:
	\begin{equation}
		W(\boldsymbol{q}, t) = A(\boldsymbol{q}, t) \hat{E}(\boldsymbol{q}, t).
	\end{equation}

	To theoretically justify W-AoI, consider any unmeasured cell $\boldsymbol{q} \notin \mathcal{S}_t$ with $A(\boldsymbol{q}, t) < A_{\max}$. Let $t_0 < t$ be its most recent reporting slot. Its AoI was reset to $t_0 - \tau(\boldsymbol{q}, t_0)$ upon reception at $t_0$. Since it remains unreported for the subsequent $t - t_0$ slots, its AoI increments by $1$ per slot, unfolding to $A(\boldsymbol{q}, t) = (t_0 - \tau(\boldsymbol{q}, t_0)) + (t - t_0) = t - \tau(\boldsymbol{q}, t_0)$. Defining $u(\boldsymbol{q}, t) \triangleq \tau(\boldsymbol{q}, t_0)$ as the generation slot of the currently stored measurement, we obtain $A(\boldsymbol{q}, t) = t - u(\boldsymbol{q}, t) \geq 1$.

	\textbf{Lemma 1.} \emph{Assume that, for every cell $\boldsymbol q\in\mathcal D$ and every slot $t$, the learning-based predictor satisfies two per-cell calibration conditions: (i) the drift that the true map accumulates since the stored measurement is bounded by the predicted AEM, i.e., $|C(\boldsymbol q,t)-C(\boldsymbol q,u(\boldsymbol q,t))|\leq \kappa_1 \hat{E}(\boldsymbol q,t) A(\boldsymbol q,t)$; and (ii) the prediction baseline error is bounded by the predicted AEM, $|C(\boldsymbol{q}, u(\boldsymbol{q}, t)) - C'(\boldsymbol{q}, t)| \leq \kappa_2 \hat{E}(\boldsymbol{q}, t)$, for constants $\kappa_1 > 0, \kappa_2 \geq 0$. Then, the true AEM of every cell is upper-bounded by the W-AoI:}
	\begin{equation}\label{eq:bound}
		E(\boldsymbol{q}, t) \leq \kappa W(\boldsymbol{q}, t), \quad \kappa = \kappa_1 + \kappa_2.
	\end{equation}
	
	\emph{Proof:} Inserting the true sampled value $C(\boldsymbol{q}, u(\boldsymbol{q}, t))$ and applying the triangle inequality yields $E(\boldsymbol{q}, t) = |C(\boldsymbol{q}, t) - C'(\boldsymbol{q}, t)| \leq |C(\boldsymbol{q}, t) - C(\boldsymbol{q}, u(\boldsymbol{q}, t))| + |C(\boldsymbol{q}, u(\boldsymbol{q}, t)) - C'(\boldsymbol{q}, t)|$, which conditions (i) and (ii) bound by $\big(\kappa_1 A(\boldsymbol{q}, t) + \kappa_2\big)\hat E(\boldsymbol q,t)$. For every cell with $A(\boldsymbol{q}, t) \geq 1$, we have $\kappa_2 \leq \kappa_2 A(\boldsymbol{q}, t)$, which establishes $E(\boldsymbol{q}, t) \leq (\kappa_1 + \kappa_2) A(\boldsymbol{q}, t) \hat{E}(\boldsymbol{q}, t) = \kappa W(\boldsymbol{q}, t)$. $\hfill \blacksquare$
	
	\emph{Remark 1:} Conditions (i) and (ii) in Lemma 1 constrain the \emph{predictor}, not the physical channel. They require the U-Net to be properly calibrated such that its announced spatial error $\hat{E}$ structurally encapsulates both the true temporal drift rate $\kappa_1$ and the inherent prediction noise $\kappa_2$.
	
	\subsection{Optimal Cell Set for AEM Reduction}
	
	By squaring and summing the bound from Lemma 1 over all cells, the total squared AEM at slot $t$ is upper-bounded by $\sum_{\boldsymbol{q} \in \mathcal{D}} E^2(\boldsymbol{q}, t) \leq \kappa^2 \sum_{\boldsymbol{q} \in \mathcal{D}} W^2(\boldsymbol{q}, t)$. 
	
	Suppose UAV $u$ selects a subset of cells $\mathcal{Q}_{u, t}$ to measure. Due to the limited uplink budget and on-board queueing, a new measurement may take $\delta \geq 0$ slots to be delivered to the central server. Upon successful delivery, the AoI of a refreshed cell resets to this delay $\delta$. Assuming that the predicted AEM $\hat{E}(\boldsymbol{q}, t)$ remains stable during this short delay, the squared AEM bound of this specific cell drops from $\kappa^2 A^2(\boldsymbol{q}, t) \hat{E}^2(\boldsymbol{q}, t)$ to $\kappa^2 \delta^2 \hat{E}^2(\boldsymbol{q}, t)$. Therefore, the total decrease in the squared AEM upper bound, which can serve as the reward function for selecting $\mathcal{Q}_{u, t}$, is defined as $
	R(\mathcal{Q}_{u, t}) = \kappa^2 \sum_{\boldsymbol{q} \in \mathcal{Q}_{u, t}} \left( A^2(\boldsymbol{q}, t) - \delta^2 \right) \hat{E}^2(\boldsymbol{q}, t).$

	Next, we consider an idealized scenario where the UAV is unrestricted by mobility and collision constraints and measurements are delivered instantly (i.e., $\delta = 0$), and obtain Proposition~1. 
	
	\textbf{Proposition 1.} \emph{For a fixed UAV $u$ with a reporting capacity of $K$ cells per slot, assume the delivery delay is zero ($\delta = 0$). To maximize the AEM reduction reward $R(\mathcal{Q}_{u, t})$, the optimal cell set $\mathcal{Q}_{u, t}^*$ consists of the $K$ cells in $\mathcal{D}$ possessing the highest W-AoI:}
	\begin{equation}\label{eq:topk}
		\mathcal{Q}_{u, t}^* = \text{TopK}_{\boldsymbol{q} \in \mathcal{D}} W(\boldsymbol{q}, t), \quad |\mathcal{Q}_{u, t}^*| \leq K.
	\end{equation}
	
	\emph{Proof:} When $\delta=0$, $R(\mathcal{Q}_{u, t})$ reduces to $\kappa^2 \sum_{\boldsymbol{q} \in \mathcal{Q}_{u, t}} W^2(\boldsymbol{q}, t)$. Under the cardinality constraint $|\mathcal{Q}_{u, t}| \leq K$, maximizing $R(\mathcal{Q}_{u, t})$ is achieved by selecting the $K$ elements with the largest $W^2(\boldsymbol{q}, t)$. Since the mapping $s \mapsto s^2$ is strictly increasing for $s \geq 0$, ranking cells by $W^2(\boldsymbol{q}, t)$ or $W(\boldsymbol{q}, t)$ induces the same Top-$K$ selection. $\hfill \blacksquare$

	\begin{figure}[h]
		\centering
		\scalebox{0.92}{\begin{tikzpicture}[font=\footnotesize, >=Latex]
				\draw[thick] (0,0) rectangle (2.9,2.9);
				\draw[densely dashed, black!45] (1.45,0) -- (1.45,2.9);
				\draw[densely dashed, black!45] (0,1.45) -- (2.9,1.45);
				\node[font=\scriptsize, black!65] at (0.28,2.72) {$\mathcal{R}_1$};
				\node[font=\scriptsize, black!65] at (2.62,2.72) {$\mathcal{R}_2$};
				\node[font=\scriptsize, black!65] at (0.28,0.17) {$\mathcal{R}_3$};
				\node[font=\scriptsize, black!65] at (2.62,0.17) {$\mathcal{R}_4$};
				\foreach \x/\y in {0.72/2.12, 2.20/2.18, 0.75/0.72, 2.16/0.66}{
					\fill[black!80] (\x,\y) circle (1.6pt);}
				\draw[->, black!55, thin] (0.72,2.12) -- (0.98,2.40);
				\draw[->, black!55, thin] (2.20,2.18) -- (1.93,2.44);
				\draw[->, black!55, thin] (0.75,0.72) -- (1.00,0.94);
				\draw[->, black!55, thin] (2.16,0.66) -- (1.91,0.92);
				\begin{scope}[shift={(6.10,1.45)}]
					\draw[densely dashed, black!55] (0,0) circle (1.85);
					\fill (0,0) circle (2.2pt);
					\node[font=\scriptsize, below left=-1pt] at (0,0) {$\boldsymbol{p}_{u,t}$};
					\node[font=\scriptsize, black!60, above=1pt] at (-0.30,1.90) {$V_{\max}\Delta t$};
					\fill[black!75] (1.20,-0.90) circle (1.5pt);
					\node[font=\scriptsize, black!75, right=1pt] at (1.20,-0.90) {8.2};
					\fill[black!75] (0.35,0.88) circle (1.5pt);
					\node[font=\scriptsize, black!75, right=1pt] at (0.35,0.88) {19.4};
					\fill[black!75] (0.30,-1.40) circle (1.5pt);
					\node[font=\scriptsize, black!75, below=1pt] at (0.30,-1.40) {22.7};
					\draw[very thick, red!72!black] (0,0) -- (-1.35,1.00);
					\foreach \t in {0.22,0.42,0.62,0.82}{
						\node[draw=red!72!black, fill=red!18, minimum size=2.0mm, inner sep=0,
						rotate=8.5] at ({-1.35*\t},{1.00*\t}) {};}
					\fill[red!72!black] (-1.35,1.00) circle (2.3pt);
					\node[font=\scriptsize, red!72!black, left=3pt, above=2pt] at (-1.35,1.00) {41.5};
					\node[font=\scriptsize, red!72!black] at (-0.62,-0.62) {measured cells};
					\draw[black!40] (-1.60,-1.55) circle (1.4pt);
					\node[font=\scriptsize, black!45, below=1pt] at (-1.60,-1.55) {unreachable};
				\end{scope}
				\node[font=\footnotesize] at (1.45,-0.62) {(a)};
				\node[font=\footnotesize] at (6.10,-0.62) {(b)};
		\end{tikzpicture}}
		\caption{Mobility-constrained trajectory. (a) Fixed Voronoi subregions, one per UAV. (b) A UAV can reach the cells within $V_{\max}\Delta t$ of $\boldsymbol{p}_{u,t}$. The numbers are weighted-AoIs and the UAV flies to the largest valid one (41.5). Cells crossed on the way are measured (shaded) but do not enter the decision.}
		\label{fig:traj}
	\end{figure}
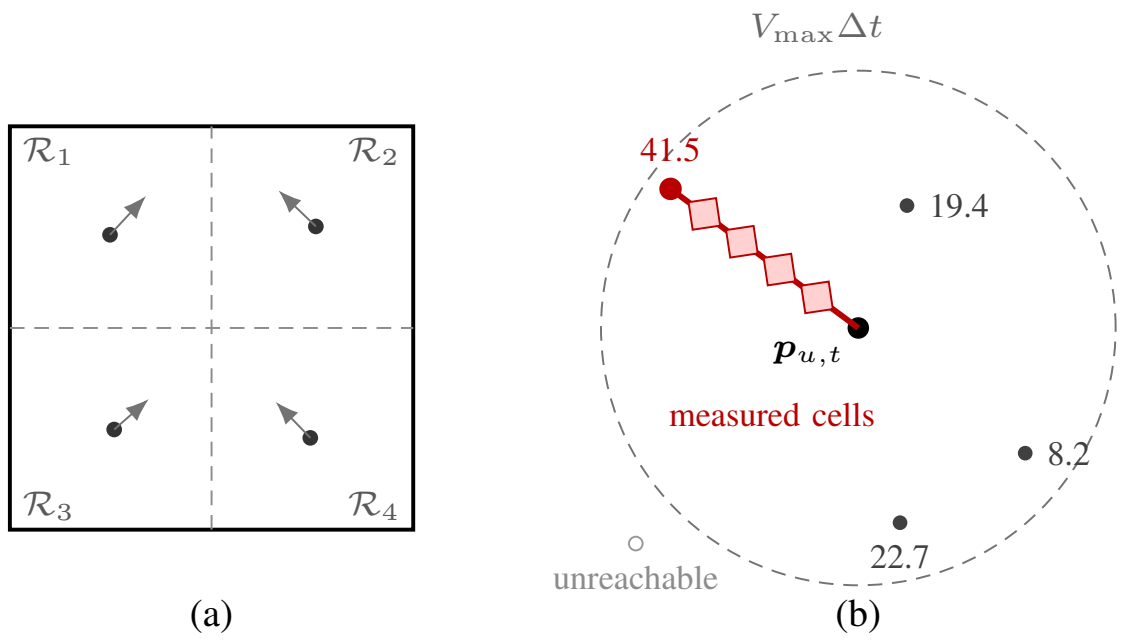
	
	\vspace{-2pt}
	\subsection{Reachable Set and Greedy Trajectory Policy}
	
	To implement the optimal Top-$K$ rule under practical mobility and collision constraints, we design a greedy trajectory policy. 
	
	To distribute the measurement workload across the fleet and prevent all UAVs from flocking to a single hotspot, the target area $\mathcal{D}$ is partitioned into $U$ disjoint subregions $\mathcal{R}_1, \dots, \mathcal{R}_U$ via a Voronoi tessellation based on the initial UAV positions $\boldsymbol{p}_{u, 0}$\footnote{If the UAVs are initially deployed in a clustered manner, a preliminary dispersion flight is executed to distribute them evenly across $\mathcal{D}$ before establishing the fixed Voronoi partitions.}. Each UAV $u$ is restricted to measure cells within its designated fixed subregion $\mathcal{R}_u$, as illustrated in Fig.~\ref{fig:traj}(a).
	
	Bounded by the maximum flight speed $V_{\max}$ and the slot duration $\Delta t$, the reachable set of cells for UAV $u$ from its current position $\boldsymbol{p}_{u, t}$ in the next slot is limited to:
	\begin{equation}
		\mathcal{Q}_{u, \text{feas}}(\boldsymbol{p}_{u, t}) = \left\{ \boldsymbol{q} \in \mathcal{R}_u : \|\boldsymbol{p}_{u, t} - \boldsymbol{q}\|_2 \leq V_{\max} \Delta t \right\}.
	\end{equation}
	
	At each slot, UAV $u$ greedily targets the reachable cell exhibiting the highest W-AoI priority:
	\begin{equation}\label{eq:greedy}
		\boldsymbol{q}_{u, t+1}^* = \arg\max_{\boldsymbol{q} \in \mathcal{Q}_{u, \text{feas}}(\boldsymbol{p}_{u, t})} W(\boldsymbol{q}, t).
	\end{equation}
	To guarantee collision-free navigation, if flying towards $\boldsymbol{q}_{u, t+1}^*$ brings UAV $u$ within the minimum safety distance $d_{\min}$ of another UAV, this candidate cell is skipped in favor of the next best cell in $\mathcal{Q}_{u, \text{feas}}$. 
	
	The selection process is visually depicted in Fig.~\ref{fig:traj}(b). It is worth noting that while the decision logic only explicitly evaluates the candidate cells, the UAV continuously senses the environment during its flight. As shown by the shaded cells in Fig. \ref{fig:traj}(b), all cells physically crossed along the straight-line trajectory to the target cell are measured. These along-path measurements form $\mathcal{P}_{u, t}$ and join the on-board queue $\mathcal{B}_u(t)$, from which \eqref{eq:report} reports the $K$ cells with the largest W-AoI, the queue-constrained counterpart of \eqref{eq:topk}, thereby prioritizing informative measurements and facilitating AEM error reduction across the map. 
	
	The complete procedure is summarized in Algorithm \ref{alg:sched}. The per-slot computational cost is dominated by the U-Net forward passes, each requiring $\mathcal{O}(N_{\text{ne}})$ operations, where $N_{\text{ne}}$ is the network size. Waypoint selection scans only the local reachable set, costing $\mathcal{O}(|\mathcal{Q}_{u,\text{feas}}|)$ per UAV, while grid traversal costs $\mathcal{O}(n_t)$ for $n_t$ crossed cells. Both costs scale with the single-slot UAV coverage area rather than the global map size.


	\begin{algorithm}[h]
		\caption{Joint Mobility and Measurement Scheduling}
		\label{alg:sched}
		\footnotesize
		\begin{algorithmic}[1]
			\Require $U$, $K$, $Q$, $T$, $a_{\rm th}$, predictors $f^{C},f^{E}$,
			$\boldsymbol{C}'(0)$, $A_0$
			\State $A\!\leftarrow\!A_0$, $\{\boldsymbol{C}'(0), \boldsymbol{Y}(0)\}\!\leftarrow\!\bar C$,
			$\mathcal{B}_u\!\leftarrow\!\emptyset$
			\For{$t=1,\dots,T$}
			\State $\boldsymbol{X}\!\leftarrow\!\big[\boldsymbol{M},\ \boldsymbol I_{t-1},\
			\boldsymbol{C}'(t\!-\!1),\ \boldsymbol Y(t\!-\!1)\big]$
			\Comment{$\boldsymbol{M}\!=\!\mathbf 1$}
			\State $\hat{\boldsymbol{C}}\!\leftarrow\!f^{C}(\boldsymbol{X})$;\ \
			$\hat{\boldsymbol{E}}\!\leftarrow\!f^{E}(\boldsymbol{X},\hat{\boldsymbol{C}})$
			\Comment{no slot-$t$ measurement}
			\State $W(\boldsymbol{q})\!\leftarrow\!A(\boldsymbol{q})\hat{E}(\boldsymbol{q})$;\ \
			$\boldsymbol{q}^{*}_{u,t+1}\!\leftarrow\!
			\arg\max_{\mathcal{Q}_{u,\mathrm{feas}}}W$
			\State Fly to $\boldsymbol{q}^{*}_{u,t+1}$ of \eqref{eq:greedy}, measure
			the crossed cells $\mathcal{P}_{u,t}$
			\State $\mathcal{B}_u(t)\!\leftarrow\!\mathcal{B}_u(t\!-\!1)\!\cup\!\mathcal{P}_{u,t}$,
			drop oldest if $|\mathcal{B}_u(t)|\!>\!Q$
			\State $\mathcal{S}_{u,t}\!\leftarrow\!\operatorname{Select}_K(\mathcal{B}_u(t))$
			of \eqref{eq:report}
			\State update $\boldsymbol C'(t)$ by \eqref{eq:update}; update $A(\boldsymbol q,t)$ by \eqref{eq:aoi}
			\State form $\boldsymbol I_t$ and the label $\boldsymbol Y(t)$ of \eqref{eq:label}
			\If{a full batch of training examples is collected}
			\State update $f^{C},f^{E}$ by one gradient step on \eqref{eq:lossc}
			\EndIf
			\EndFor
		\end{algorithmic}
	\end{algorithm}

	\section{Performance Evaluation}
	\label{sec:exp}
	\subsection{Scenario and Parameters}
	We simulate a $300\times300$~m area on a 2-D grid with
	$\Delta_D=5$~m ($I=J=60$), UAV altitude $h_0=100$~m, and
	$M=6$ GBSs. Each episode lasts $1000$ slots ($\Delta t=1$~s).
	The predictors are trained for $100$ episodes ($10^5$ slots) and
	evaluated over $10$ unseen seeds; results are averaged over the last
	$100$ slots of each evaluation episode to exclude the initial transient.
	The area contains eight non-overlapping buildings with side lengths
	uniformly drawn from $[50,70]$~m and truncated-Rayleigh heights
	(mean $30$~m), covering about $30\%$ of the area. Building blockage
	determines each GBS--cell link's LoS/NLoS branch, while UAVs fly above
	all rooftops. Each GBS uses an eight-element vertical array with
	$-10^\circ$ downtilt \cite{zhang2020radio}.
	The shared band is $180$~kHz at $2$~GHz, with load factors
	$\{0.05,0.95,0.10,0.90,0.05,0.95\}$.
	The maximum flight distance is $30$~m per slot, corresponding to
	$n_{\max}=10$ crossed cells (seven for an axis-aligned flight), and
	$d_{\min}=10$~m. Unless otherwise stated, $U=4$, $K=4$, and $Q=32$,
	so $KU=16$ cells can be refreshed per slot and
	$a_{\rm th}=\lceil IJ/(KU)\rceil=225$.
	The remaining parameters are listed in Table~\ref{tab:param}.
	
	\begin{table}[t]
		\caption{Simulation Parameters}
		\label{tab:param}
		\centering
		\footnotesize
		\setlength{\tabcolsep}{2.5pt}
		\begin{tabular*}{\columnwidth}{@{\extracolsep{\fill}}l c l c}
			\toprule
			\textbf{Parameter} & \textbf{Value} & \textbf{Parameter} & \textbf{Value}\\
			\midrule
			$V_{\max}$ & $30$ m/s & $A_0$ & $10$\\
			Queue $Q$ & $32$ & $A_{\max}$ & $2000$\\
			Tx power & $24$ dBm & Noise PSD & $-169$ dBm/Hz\\
			Noise fig. & $9$ dB & $\rho_h$/$\rho_\alpha$ & $0.99$\\
			Params (each) & $1.2$ M & $\lambda_1$/$\lambda_2$ & $4$/$0.4$\\
			Optimizer & Adam & $N_{\rm ma}$/$p$ & $8$/$0.8$\\
			U-net base ch. & $24$ & Batch & $32$\\
			lr (train/eval) & $10^{-3}$/$10^{-4}$ & Episodes (train/eval) & $100$/$10$\\
			\bottomrule
		\end{tabular*}
	\end{table}
	
	\subsection{Baselines and Metrics}
	All policies share the same channel realizations, initial maps and
	UAV positions, Voronoi subregions, measurement budgets, and random
	seeds. Random selects reachable waypoints and reports randomly;
	FIFO follows a serpentine route and reports the oldest queued samples.
	AoI-only ($W=A$) and AEM-only ($W=\hat E$) use one factor to rank
	both waypoints and reports, while W-AoI uses their product.
	Oracle replaces $\hat E$ with the true AEM $E$ and serves only as a
	reference, since the trajectory policy remains greedy.
	Kriging \cite{sato2017kriging} and Diffusion \cite{rong20253dckmdiff} use the same waypoint and report selection rules but replace the proposed temporal CKM/AEM predictor. Kriging performs online interpolation using measurements with $A\le a_{\rm th}$, whereas Diffusion adopts the same 2-D U-Net backbone with iterative denoising and online training. All predictors are trained and evaluated on the same hardware platform under identical software configurations.

	\subsection{Results}
	Table~\ref{tab:res} reports the steady-state MSE, its reduction relative to the static map $\bar C$ (Gain), and the amortized per-map computational time (Time), including online training overhead. The pure inference times are $3.06$~ms for 2D-U-Net and $56.9$~ms for Diffusion. The amortized times in Table~\ref{tab:res} are $8$~ms for the proposed predictor, compared with $39.4$~ms for Kriging and $145.6$~ms for Diffusion, demonstrating its clear computational advantage. W-AoI achieves $5.51$~dB$^2$, reducing the
	error by $3.77$ and $4.26$~dB$^2$ over AoI-only ($9.28$) and
	AEM-only ($9.77$), and by $4.82$ and $5.01$~dB$^2$ over FIFO
	($10.33$) and Random ($10.52$), respectively. Neither freshness nor predicted error alone captures both sources of estimation error. Oracle obtains $5.06$~dB$^2$, only $0.45$~dB$^2$ below W-AoI. A two-sided paired $t$-test across the ten episodes gives $p<0.05$ against every baseline and $p\approx0.56$ against Oracle, indicating significant gains over the baselines but no significant gap from Oracle.
	
	\begin{table}[h]
		\centering
		\caption{Steady-state estimation error at $K=4$, $U=4$; mean $\pm$ std over the
			cells and the $10$ evaluation episodes.}
		\label{tab:res}
		\footnotesize
		\setlength{\tabcolsep}{2.7pt}
		\begin{tabular}{@{}llccc@{}}
			\toprule
			Policy & Constructor & MSE (dB$^{2}$) & Gain (dB$^{2}$) & Time (ms)\\
			\midrule
			Random & 2-D U-Net & $10.52\pm6.01$ & $4.47\pm6.43$ & $\mathbf{8}$\\
			FIFO & 2-D U-Net & $10.33\pm5.07$ & $4.66\pm5.39$ & $\mathbf{8}$\\
			AoI-only & 2-D U-Net & $9.28\pm5.49$ & $5.72\pm4.97$ & $\mathbf{8}$\\
			AEM-only & 2-D U-Net & $9.77\pm5.59$ & $5.22\pm4.86$ & $\mathbf{8}$\\
			\midrule
			\textbf{W-AoI (Proposed)} & 2-D U-Net & $\mathbf{5.51\pm4.70}$ & $\mathbf{9.49\pm5.22}$ & $\mathbf{8}$\\
			\textbf{Oracle} & 2-D U-Net & $\mathbf{5.06\pm4.89}$ & $\mathbf{9.94\pm5.42}$ & $\mathbf{8}$\\
			\midrule
			W-AoI & Kriging \cite{sato2017kriging} & $11.49\pm4.77$ & $3.51\pm4.86$ & $39.4$\\
			W-AoI & Diffusion \cite{rong20253dckmdiff} & $13.21\pm3.33$ & $1.79\pm2.66$ & $145.6$\\
			No refresh & Static $\bar C$ & $15.00\pm1.87$ & $0$ & $0$\\
			\bottomrule
		\end{tabular}
	\end{table}
	
	Replacing the temporal predictor with Kriging or Diffusion increases
	the MSE to $11.49$ and $13.21$~dB$^2$, respectively, both worse than
	Random. Kriging treats measurements collected up to $a_{\rm th}$
	slots apart as samples of one stationary map, thereby smoothing
	temporal drift; the temporal predictor instead uses the previous
	reconstruction to track this drift. Diffusion generates plausible
	maps through iterative denoising, whereas the cell-wise MSE favors
	direct regression and is affected by generation variance. 
	
	\begin{figure}[h]
		\centering
		\includegraphics[width=\columnwidth,trim=0 7 0 0,clip]{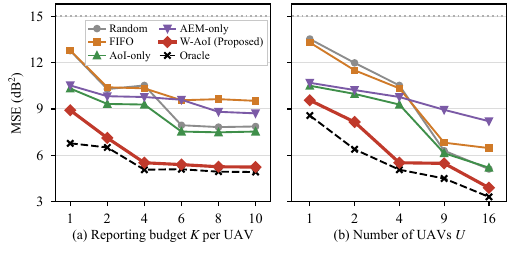}
		\caption{Estimation error against the per-UAV reporting budget $K$ at $U=4$ and the
			number of UAVs $U$ at $K=4$. Each point averages the last $100$ slots of the $10$
			evaluation episodes; the dotted line is the MSE of the static reference map $\bar C$}.
		\label{fig:sweep}
	\end{figure}
	
	\noindent\textbf{Effect of reporting budget $K$.}\label{sec:K}
	Fig.~\ref{fig:sweep}~(a) evaluates
	$K\in\{1,2,4,6,8,10\}$ at $U=4$.
	At $K=1$, only four measurements are reported per slot although
    a flight can cross ten cells; the queue therefore grows and
	the proposed rule improves over AoI-only by just $1.41$~dB$^2$
	($8.91$ versus $10.32$). Increasing $K$ from $2$ to $4$
	reduces the MSE by $1.61$~dB$^2$; beyond $K=4$, it changes
	by less than $0.3$~dB$^2$ ($5.51$, $5.39$, $5.25$, and
	$5.23$ for $K=4,6,8,10$). Once the reporting budget can
	accommodate the cells crossed during flight, the queue is cleared
	within a slot and mobility, rather than reporting capacity,
	determines freshness. For $K\ge2$, the gain over AoI-only remains
	$2.1$--$3.8$~dB$^2$, while the Oracle gap narrows from
	$0.62$~dB$^2$ at $K=2$ to about $0.3$~dB$^2$ for $K\ge6$,
	showing that prediction error becomes the main remaining limitation.

	\noindent\textbf{Effect of fleet size $U$.}\label{sec:U}
	Fig.~\ref{fig:sweep}~(b) evaluates
	$U\in\{1,2,4,9,16\}$ at $K=4$.
	The fleet refreshes $KU$ cells per slot, giving a mean revisit
	interval of $IJ/(KU)$, from $900$ slots at $U=1$ to $56$
	at $U=16$. At $U=1$, all policies have MSE above
	$8.5$~dB$^2$, and W-AoI improves over AoI-only by
	$0.94$~dB$^2$. Its gain peaks at $U=4$ ($3.77$~dB$^2$),
	where the revisit interval is about twice the fading correlation
	time, $1/(1-\rho_h)\approx100$ slots: channel changes are
	large enough for predicted errors to distinguish cells, while
	their ages remain relevant to scheduling. With fewer UAVs,
	most cells become similarly stale; with more UAVs, fewer cells
	remain stale. For $U>9$, the revisit interval falls below the
	correlation time and all policies improve. The gap between Random
	and W-AoI narrows from $5.0$~dB$^2$ at $U=4$ to
	$0.82$ and $1.21$~dB$^2$ at $U=9$ and $16$, respectively.
	At $U=16$, W-AoI achieves $3.89$~dB$^2$, outperforming
	the $U=4$ Oracle ($5.06$~dB$^2$).

	\section{Conclusion}
	This paper studied the long-term CKM estimation error minimization problem, to properly maintain the dynamic CKM. We proposed the W-AoI driven UAV scheduling approach to tackle the unobservable dynamic errors. We proved that by fusing exact temporal staleness with U-Net-predicted spatial uncertainty, the W-AoI metric provides a theoretically guaranteed upper bound on the reconstruction error. We proposed the lightweight greedy trajectory policy incorporating continuous along-path sensing, which maximizes data yield to accelerate map updates. Extensive simulations validated the superiority of our proposed method.

	\bibliographystyle{IEEEtran}
	\bibliography{references}
	
\end{document}